\pdfoutput=1 
 
\documentclass[acmtog,nonacm]{acmart}

\usepackage{booktabs} 

\definecolor{darkorange}{rgb}{1.0, 0.55, 0.0}

\usepackage[ruled]{algorithm2e} 

\SetAlFnt{\small}
\SetAlCapFnt{\small}
\SetAlCapNameFnt{\small}
\SetAlCapHSkip{0pt}

\usepackage{amsmath}
\usepackage{bm}

\acmJournal{TOG}

\renewcommand\footnotetextcopyrightpermission[1]{} 
 \setcopyright{none}
\begin{document}

\title{Style and Pose Control for Image Synthesis of Humans from a Single Monocular View}

\author{KRIPASINDHU SARKAR}
\affiliation{\institution{Max Planck Institute for Informatics}
 }
 \email{ksarkar@mpi-inf.mpg.de}
 \author{VLADISLAV GOLYANIK}
\affiliation{\institution{Max Planck Institute for Informatics}
 }
 \email{golyanik@mpi-inf.mpg.de}
\author{LINGJIE LIU}
\affiliation{\institution{Max Planck Institute for Informatics}
 }
 \email{lliu@mpi-inf.mpg.de}
\author{CHRISTIAN THEOBALT }
\affiliation{\institution{Max Planck Institute for Informatics}
 }
 \email{theobalt@mpi-inf.mpg.de}

\renewcommand\shortauthors{Sarkar, K. et al}

\begin{abstract} 
Photo-realistic re-rendering of a human from a single image with explicit control over body pose, shape and appearance enables a wide range of applications, such as human appearance transfer, virtual try-on, motion imitation, and novel view synthesis.
While significant progress has been made in this direction using learning-based image generation tools, such as GANs, existing approaches yield noticeable artefacts such as blurring of fine details, unrealistic distortions of the body parts and garments as well as severe changes of the textures. 
We, therefore, propose a new method for synthesising photo-realistic human images with explicit control over pose and part-based appearance, \textit{i.e.,}  StylePoseGAN, where we extend a non-controllable generator to accept conditioning of pose and appearance separately. 
Our network can be trained in a fully supervised way with human images to disentangle pose, appearance and body parts, and it 
significantly outperforms existing single image re-rendering methods.
Our disentangled representation opens up further applications such as garment transfer, motion transfer, virtual try-on, head (identity) swap and appearance interpolation. 
StylePoseGAN achieves state-of-the-art image generation fidelity on common perceptual metrics compared to the current best-performing methods and convinces in a comprehensive user study. 

\end{abstract}

%
%


%
%


\begin{teaserfigure}
  \centering
   \includegraphics[width=.95\textwidth]{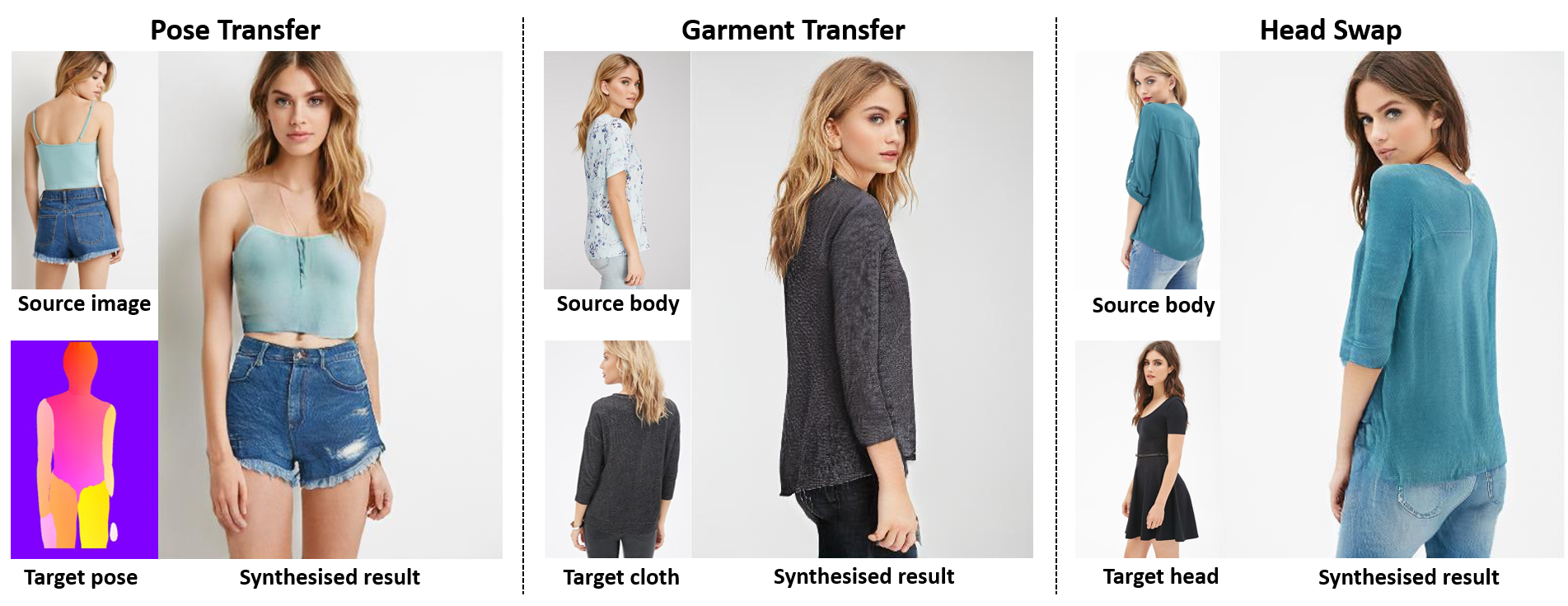}
   \caption{\textbf{We present StylePoseGAN, \textit{i.e.,} a new approach for synthesising photo-realistic novel views of a human from a single input image with explicit control over pose and per-body-part appearance.} 
   We generate images of higher fidelity compared to the state-of-the-art methods, especially with fine appearance details such as faces and texture patterns. 
   Our method enables several applications such as pose transfer, garment transfer, and head swap. 
   All synthesised results in this figure are shown as obtained from our model without further post-processing. 
   }
   \label{fig:teaser}
\end{teaserfigure}

\maketitle
\pagestyle{plain}

\section{Introduction} 
\label{sec:intro} 

Creating photo-realistic images and videos of humans under full control of pose,  shape and appearance is a core challenge in computer animation with many  applications in movie production, content creation, visual effects and virtual  reality, among others.
Achieving this with the established computer graphics toolchains is an extremely complex and time-consuming process. First, a high-quality 3D human geometry and appearance model is required either manually designed by skilled artists or captured with dense camera arrays. To make the model animatable, sophisticated rigging techniques need to apply, and some manual post-processing (\textit{e.g.,} skinning weight painting) is often required. After that, computationally-expensive global illumination rendering techniques are needed to render the model photo-realistically. Some works \cite{Xu:SIGGRPAH:2011,casas14,Volino2014} attempted to avoid such sophisticated toolchains with image-based rendering (IBR) techniques. However, these methods still have sub-optimal rendering quality and limited control over the renderings and are often scene-specific. 

Recently, great progress in learning-based approaches for image synthesis from a single monocular input view has been made \cite{Neverova2018, esser2018variational,  Pumarola_2018_CVPR,SiaroSLS2017,KratzHPV2017, Sarkar2020,Grigorev2019CoordinateBasedTI,yoon2020poseguided,alldieck2019tex2shape,Lazova2019360DegreeTO}. 
However, because of the underconstrained nature of this problem, most methods have limits in the visual quality of the results. 
They frequently exhibit oversmoothing, add unrealistic textures, lack fine-scaled details and wrinkle patterns, and lose facial identity.
This paper proposes a new lightweight learning-based approach for photo-realistic controllable human image generation, \textit{i.e.,} StylePoseGAN, that enables fine-grained control over the generated human images and significantly outperforms all tested competing methods for re-rendering of humans from a single image, see Fig.~\ref{fig:teaser} for the method overview. 
Our approach works with a single camera only, and no expensive set-ups and manual user interaction are required. 
At its core is a conditional generative model with several lightweight proxies, enabling control over body pose and appearance from a single image. 

\begin{figure}[t!]
    \includegraphics[width=1.0\linewidth]{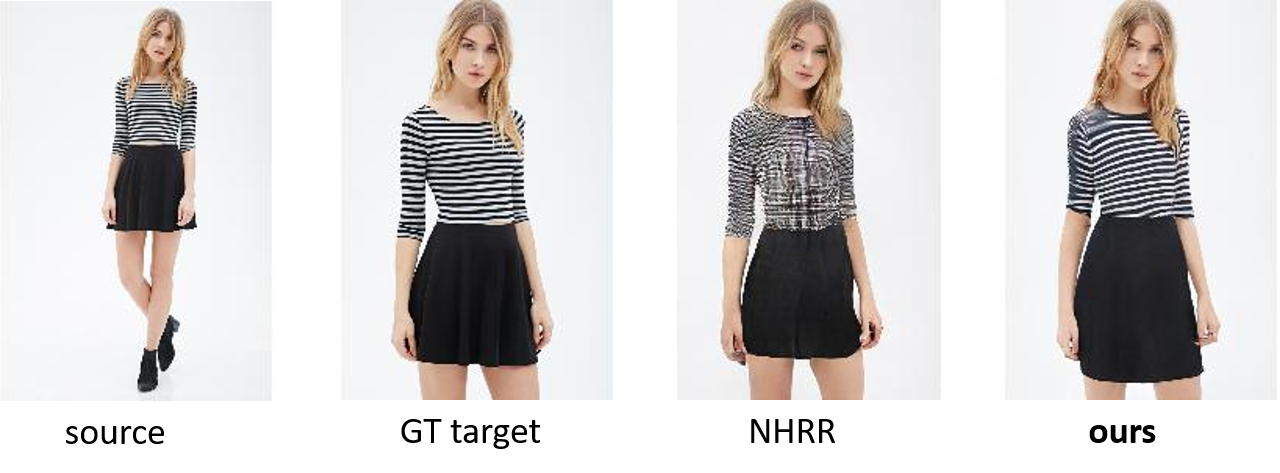}
    \caption{Most of the recent human re-rendering methods cannot recover the fine texture details. 
    Here we show the result of the recent state-of-the-art method of NHRR \cite{Sarkar2020} and compare it with ours. 
    }
    \label{fig:comparison_NHRR} 
\end{figure}

Our approach is an extension of non-controllable\footnote{not controllable for style and pose on the level of traditional graphics pipelines} and the state-of-the-art photo-realistic generative models for human portraits such as StyleGAN  \cite{karras2019style,  Karras2019stylegan2}. 
%
StylePoseGAN improves upon it by conditioning pose as spatial features and conditioning appearance for the weight demodulation. Our model can be trained in a fully supervised way with human images that enables the reconstruction of images with a forward pass. 
Such an approach results in significantly fewer artefacts and better preservation of fine texture patterns, compared to the results of existing methods (Sec.~\ref{sec:results}). 
%
Due to the separate conditioning of two aspects in different modalities, our design explicitly disentangles pose, appearance and body parts, and opens up several applications such as pose transfer, garment transfer, motion transfer and parts-swap.
Compared to state of the art, we support much finer texture details and richer texture patterns of the garments, which is one of the strongest properties of StylePoseGAN, see  Fig.~\ref{fig:comparison_NHRR} for qualitative comparison with NHRR \cite{Sarkar2020}. 
To summarise, our technical \textbf{contributions} are: 
\begin{itemize} 
 \item StylePoseGAN, \textit{i.e.,} a disentangled architecture for human image generation  which pushes the attainable visual quality of the generated human images significantly. 
 \item State-of-the-art results on the DeepFashion dataset \cite{Liu2016DeepFashion} which are confirmed with quantitative metrics, and qualitatively with a user study. 
 \item Support of richer texture and pose-dependent effects such as wrinkles and shadings, compared to the current state of the art. 
 \item Applications of StylePoseGAN in image manipulation, \textit{i.e.,} garment transfer, head (identity swap) and image interpolation, which enable special effects which can be useful in computer animation. 
\end{itemize} 

We evaluate our method on several datasets and report perceptual metrics. 
Additionally, we conduct a comprehensive user study. 
We encourage the readers to watch our supplementary video. 

\section{Related Work}

We next review related works in human rendering, deep  generative models and human pose transfer. 

\subsection{Classical and Neural Rendering of  Humans} 
Photo-realistic rendering of a real human using classical rendering methods heavily relies on a high-quality geometry and appearance human model. To achieve high quality, real individuals need to be captured with sophisticated scanners and reconstructed by 3D reconstruction techniques. 
To synthesise the human image in a new pose, some sophisticated rigging techniques need to be applied to rig the human model, and a mapping from body poses to pose-dependent appearance or geometry models need to be learned. 
\cite{Xu:SIGGRPAH:2011} propose a method which first retrieves the most similar poses and view-points in a pre-captured database and then applies retrieval-based texture synthesis. 
\cite{casas14,Volino2014} compute a temporally coherent layered representation of appearance in texture space. 
However, the synthesis quality is limited by the quality of the geometry and appearance human model. 
To address the limitations of classical rendering methods, recent works integrated deep learning techniques into the classical rendering pipelines. 
Some methods~\cite{Thies2019,Kim2018,Liu2019,liu2020NeuralHumanRendering,Meshry2019,yoon2020poseguided,kappel2020high-fidelity} first render 2D/3D skeleton, 2D joint heat maps, or a coarse surface geometry with explicit or neural textures into coarse RGB images or feature maps which are then translated into high-quality images using image translation networks, such as pix2pix~\cite{pix2pix2017}. 
Another line of works learn scene representations for novel view synthesis from 2D images. 
Although this kind of methods achieves impressive renderings of static \cite{sitzmann2019deepvoxels,Sitzmann2019,Mildenhall20eccv_nerf,Liu20neurips_sparse_nerf,Zhang20arxiv_nerf++} and dynamic scenes and enables playback and  interpolation~\cite{lombardi2019neural,Zhang20arxiv_nerf++,Park20arxiv_nerfies,Pumarola20arxiv_D_NeRF,Tretschk20arxiv_NR-NeRF,Li20arxiv_nsff,Xian20arxiv_stnif,peng2020neural,raj2020pva,wang2020learning, Gafni20arxiv_DNRF}, it is not straightforward to extend these methods to synthesise human images of the full body with explicit control. 
%
%
Moreover, most of them are scene-specific. 
In contrast, our StylePoseGAN builds upon deep generative models and can synthesise photo-realistic human images with explicit control over body pose and human appearance.

\subsection{Deep Generative Models} 

Generative Adversarial Networks (GAN) have made remarkable achievements in image generation in recent years. 
A GAN model has two components: generator and discriminator. 
The core idea of a GAN model is to use a generator to synthesise a candidate image from a noise vector $\mathbf{z}$ sampled from the distribution of training images and let the discriminator evaluate whether the candidate is real or fake. 
The two components are trained together until the image generated by the generator is realistic enough to fool the discriminator. 
The first GAN model was introduced by Goodfellow et al.~\shortcite{Goodfellow14}, which was only able to synthesise low-resolution images with limited quality. 
To improve the quality, 
SAGAN~\cite{sagan} introduces a self-attention mechanism into convolutional GANs, which allows the generator synthesising details using cues from all feature locations and the discriminator checking features in global image space. 
BigGAN \cite{Brock2018Biggan} makes multiple changes on SAGAN for  quality improvement.
ProGAN~\cite{karras2018progressive} demonstrates photo-realistic images of human faces in a high resolution of $1024 \times 1024$ by training the generator and discriminator progressively from low resolution to high resolution.
Since these methods use a single latent vector $\mathbf{z}$ to resemble the latent distribution of training data, they cannot  disentangle different attributes in the images so have limited control over image synthesis. 
StyleGANs~\cite{karras2019style, Karras2019stylegan2} 
approach this problem by mapping $\mathbf{z}$ to an intermediate latent space $\mathbf{w}$, which is then fed into the generator to control different levels of attributes. 
Although they provide 
more control on image synthesis, it is still not able to completely disentangle different semantically meaningful attributes and control them in the synthesis. 
%
Recent works~\cite{tewari2020stylerig,tewari2020pie} extend StyleGAN to synthesise face images with a rig-like control over 3D interpretable face parameters such as face pose, expressions and scene illumination. 
GAN-control~\cite{shoshan2021gancontrol} employ contrasting learning to train GANs with an explicitly disentangled latent space for faces, which can control identity, age, pose, expression, hair colour and illumination. 
%
Compared to faces, synthesising the full human appearance with control of 3D body pose and human appearance is a much more difficult problem due to more severe 3D pose and appearance changes. 
%

Conditional GAN (cGAN) is a type of GAN, which provides conditional information for the generator and discriminator. cGAN is useful for applications such as class conditional image generation~\cite{mirza2014conditional,miyato2018cgans,pmlr-v70-odena17a} and image to image translation~\cite{pix2pix2017,wang2018pix2pixHD}. 
Most works~\cite{mirza2014conditional,pix2pix2017,wang2018pix2pixHD,park2019SPADE,wang2018vid2vid} require paired data for fully-supervised training. 
pix2pix~\cite{pix2pix2017} and pix2pixHD~\cite{wang2018pix2pixHD} learn the mapping from input images to output images. 
GauGAN~\cite{park2019SPADE} focuses on image generation from segmentation masks and designs an interactive tool for users to control over semantic and style in the image synthesis. 
To tackle the setting that paired data are unavailable, some works~\cite{CycleGAN2017,YiZTG2017,LiuBK2017,choi2017stargan,Recycle-GAN} learns the mapping between two domains based on unpaired data. 
CycleGAN~\cite{CycleGAN2017} introduces a cycle consistency loss to enforce the translation property, \textit{i.e.,} that the inverse mapping of a mapping of an image should result in the original image. 
%
We propose a method that extends StyleGANs \cite{karras2019style, Karras2019stylegan2} and can synthesise  photo-realistic images of a full human body with explicit control over 3D poses and the appearance of each body part. 
A concurrent work~\cite{lewis2020vogue} is similar to our method, which proposed a pose-conditioned StyleGAN2 latent space interpolation for virtual try-on. In contrast to their method, 1) we represent appearance in a pose independent normalised space which makes part based conditioning easier, 2) we use an explicit appearance encoder to encode the part based appearance latent vector $z$ with a single forward pass instead of optimizing for $z$, 3) we do not use explicit supervision through segmentation masks as needed in~\cite{lewis2020vogue} for latent code optimisation.

\subsection{Human Pose Transfer} 
The human pose transfer problem is defined as transferring person appearance from one pose to another \cite{MaSJSTV2017}. 
Most approaches formulate it as an image-to-image mapping problem, \textit{i.e.,} given a reference image of the target person, mapping the body pose in the format of renderings of a skeleton~\cite{chan2019dance,SiaroSLS2017,Pumarola_2018_CVPR,KratzHPV2017,zhu2019progressive}, dense  mesh~\cite{Liu2019,wang2018vid2vid,liu2020NeuralHumanRendering,Sarkar2020,Neverova2018,Grigorev2019CoordinateBasedTI,yoon2020poseguided,kappel2020high-fidelity} or joint position heatmaps~\cite{MaSJSTV2017,Lischinski2018,Ma18} to real images. 
Ma et al.~\shortcite{MaSJSTV2017} design a two-stage framework, which first generates a coarse image of the person in the reference image with the target pose and refines the coarse image with a UNet trained in an adversarial way. 
To better preserve the appearance from the reference image to the generated image, some methods~\cite{liu2020NeuralHumanRendering,Sarkar2020} first map the human appearance in the screen space to UV space and feed the rendering of the person in the target pose with the UV texture map into an image-to-image translation network. 
Thanks to the explicit control over the pose and per-body-part appearance, StylePoseGAN can perform not only pose transfer but also be used for garment transfer and identity exchange. 

\section{Method}
\label{sec:method}

\begin{figure*}[t!]
    \includegraphics[width=\linewidth]{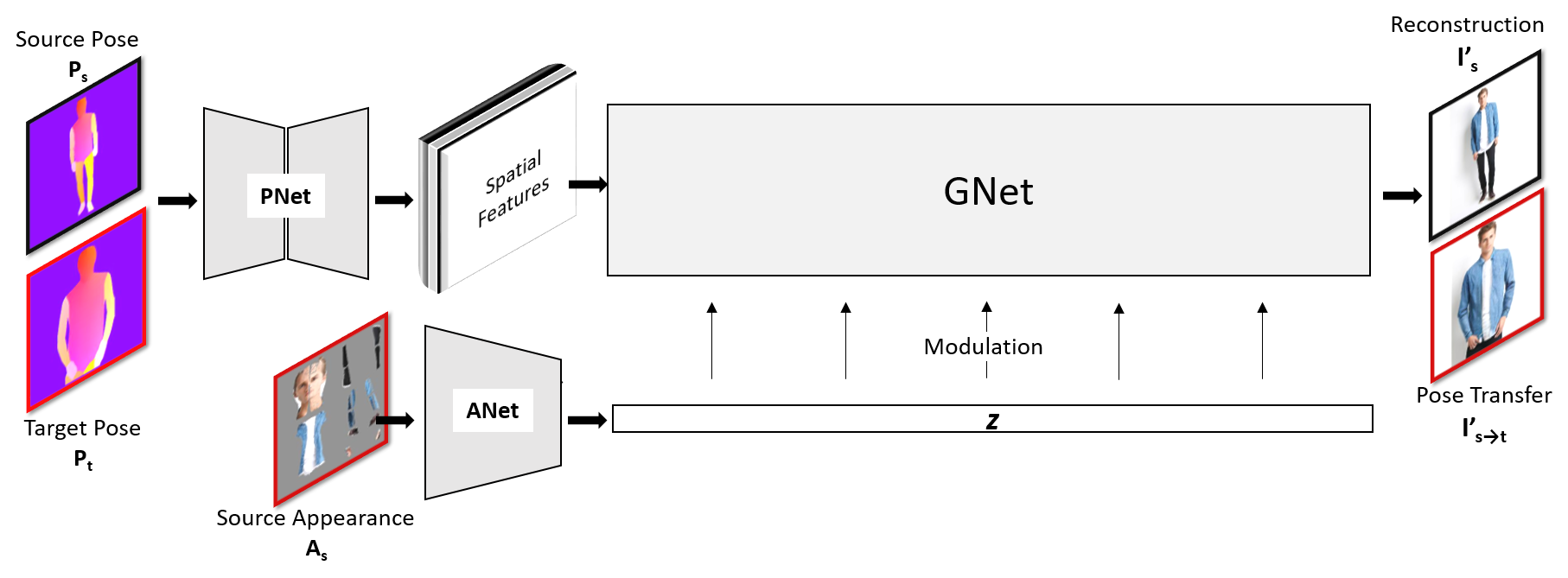}
    \caption{\textbf{Overview of StylePoseGAN}: Given an image $I$ of a person, we extract the  pose $P$ and appearance $A$ using DensePose \cite{Guler2018DensePose}. 
    We then encode the pose and appearance to encodings $E$ and $z$ 
    such that $E$ is a tensor and $z$ is a vector. 
    We finally condition a high-fidelity style-based generator with the extracted pose and appearance to reconstruct back $I$. 
    The pose and appearance are further disentangled by training with image pairs $(I_s, I_t)$ of the same person with a different pose where we use the appearance of source $A_s$ and the pose of the target $P_t$ to reconstruct $I_t$. 
    The entire pipeline is trained end-to-end in a fully supervised manner with image reconstruction loss and adversarial loss. 
}
    \label{fig:pipeline} 
\end{figure*}

Given a single image $I$ of a person, our goal is to synthesise a new image of the same person in a different target body pose. 
The overall idea can be summarised as follows. 
We first extract pose $P$ and appearance $A$ from $I$. 
Second, we encode pose and appearance to the tensor encoding $E$ and $z$, respectively where $E$ is a 3D tensor with spatial dimensions (height and width), whereas $z$ is a vector.
We then reconstruct $I$ using a high-fidelity style-based generator with $z$ as the noise vector and $E$ as the spatial input. 
The pose and appearance are further disentangled by training with the source ($s$) -- target ($t$) image pairs $(I_s, I_t)$ of the same person with different poses, where we use the appearance of source $A_s$ and the pose of the target $P_t$ to reconstruct $I_t$ in a fully supervised manner. 
Our method is summarised in Fig.~\ref{fig:pipeline}, and is described in detail in the following. 

\begin{figure}[t!]
    \includegraphics[width=\linewidth]{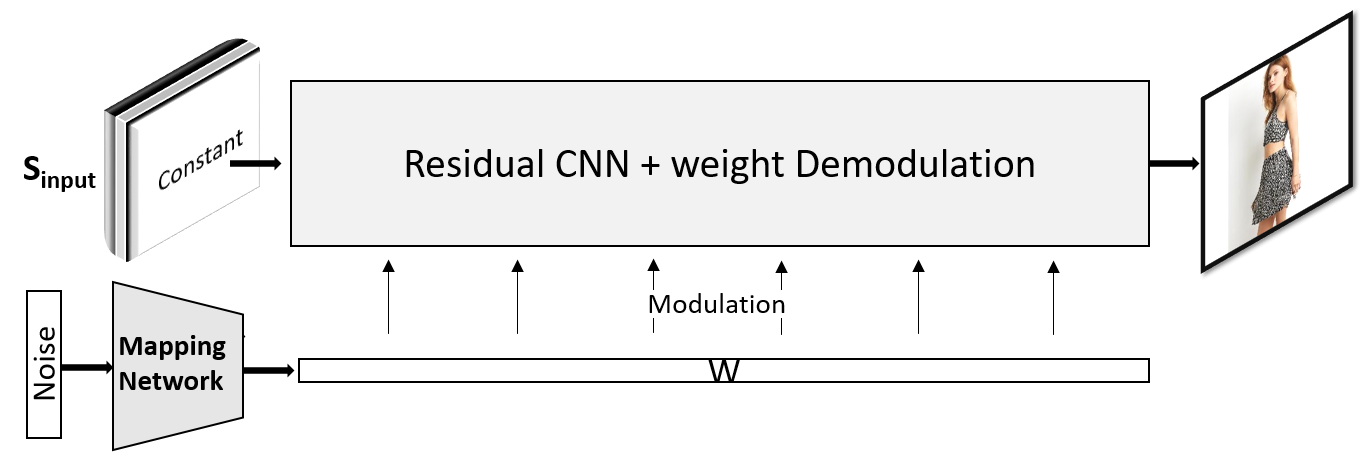}
    \caption{\textbf{The original StyleGAN2 architecture \cite{Karras2019stylegan2}.} StyleGAN-based methods take a constant 3D tensor with spatial dimensions (with a predefined \textit{height} $\times$ \textit{width} $\times$ \textit{channel})
    as input which is translated to the three-channel output image by series of \textit{demodulated} convolutions controlled by the learned latent vector $W$. 
    In StylePoseGAN, we use the tensor $S_{input}$ to provide spatial conditioning, while $W$ is used for appearance conditioning. 
    See Fig.~\ref{fig:pipeline} for an overview of our method and further details. 
    }
    \label{fig:stylegan2} 
\end{figure}

\subsection{Pose and Appearance  Extraction}\label{ssec:pose_and_appearance_extraction} 

We use DensePose~\cite{Guler2018DensePose} to detect the human pose $P \in \mathbb{R}^{H \times W \times 3}$ from the image $I$, and represent the appearance $A \in \mathbb{R}^{H_a \times W_a \times 3}$ with the partial texture map of the underlying SMPL mesh. 
Here, $H \times W$ is the resolution of the generated image (with $H$ and $W$ denoting its height and width, respectively), and $H_a \times W_a$ is the resolution of the texture image (with $H_a$ and $W_a$ denoting its height and width, respectively). 
DensePose provides pixelwise correspondences between a human image and the texture map of a parametric model of human SMPL \cite{SMPL:2015}. 
This makes the computation of the partial texture from the visible body parts in the human image possible with a single sampling operation. 
The ResNet-101-based DensePose network provided by the authors is pre-trained on COCO-DensePose dataset, and outputs $24$ part-specific \textit{U, V} coordinates of the SMPL model. 
For easier mapping, the $24$ part-specific UV-maps are combined to a single UV-texture map $A$ in the format provided in the SURREAL dataset \cite{varol17_surreal} through a pre-computed lookup table. 

\begin{figure}[t!]
    \includegraphics[width=\linewidth]{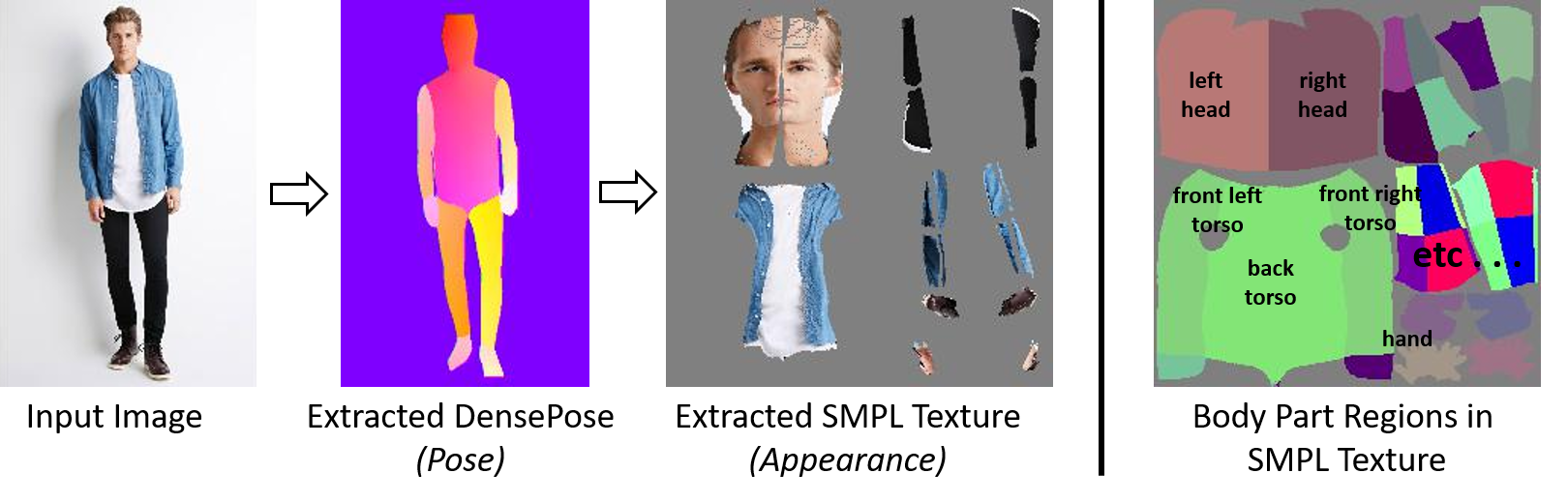}
    \caption{\textbf{Pose and Appearance Extraction.} We use
    partial texture map of the SMPL mesh as the appearance $A$ and 
    DensePose to represent the pose $P$ of the human. 
    The partial texture image $A$ is divided into $24$ part-specific regions $p_i$ (\textit{e.g.,} left head, right head, upper left arm, \textit{etc.}) as shown on the right. 
    We can provide part-specific conditioning by changing $A$ in the corresponding regions. 
    }
    \label{fig:appearance_ext} 
\end{figure}

The normalised texture map provides a pose-independent appearance encoding of the subject, where each part (out of $24$) is assigned a specific region or a set of pixels in $A$. 
Therefore, we can perform part-specific conditioning for the parts 
$p_i$, $i \in \{1, \hdots, 24\}$
by changing values at its corresponding region $A[\mathcal{S}_{p_i}]$. 
Here, $\mathcal{S}_{p_i} \subset \mathbb{R}^{H_a \times W_a}$ are the pixel locations of the part $p_i$ in the normalised texture map, and  $A[\cdot]$ is the indexing operation.
See Fig.~\ref{fig:appearance_ext}, \ref{fig:hybrid_appearance} and Sec.~\ref{sec:garment_transfer} for more details. 

\subsection{Pose and Appearance Encoding} 

\paragraph{PNet} We encode $P$ by a fully convolutional network \textit{PNet} comprising four downsampling residual blocks that produce the encoded pose $E \in \mathbb{R}^{H/16 \; \times \; W/16 \; \times \; 512}$. 
Note that based on the design of the generator that is used subsequently in our architecture, we do not need an explicit pose encoder. 
However, an encoded tensor with smaller spatial dimension is more suited for the StyleGAN2 based generator. 
See Secs.~\ref{sec:generator} and \ref{sec:ablation} for more details. 

\paragraph{ANet} The appearance encoder \textit{ANet} has the same architecture as \textit{PNet} in its initial part. 
In contrast to \textit{PNet}, its spatial activation volume is further passed through convolutional layers and, finally, a fully connected layer to produce the appearance encoding $z \in \mathbb{R}^{2048}$. 
Despite of the common architecture in the initial layers,  \textit{ANet} and \textit{PNet} do not share any weights. 

\subsection{Image Generation with a Style-based  Generator}\label{sec:generator} 
We use a StyleGAN2-based generator \cite{Karras2019stylegan2}  \textit{GNet} that combines the pose encoding $E$ and the appearance encoding $z$ to reconstruct back the input image $I = \text{\textit{GNet}}(E, z)$. 
The original StyleGAN takes a constant tensor $S_{input}$ with spatial dimensions (with a predefined \textit{height} $\times$ \textit{width} $\times$ \textit{channel}) as input on which convolutions are performed. 
A separate latent noise vector that controls the generated image is passed through a mapping network, and its output $w$ is used to modulate the weights of the convolution layers, see Fig.~\ref{fig:stylegan2}. 
We observe that the tensor $S_{input}$ can be used to provide spatial condition to the generator, instead of keeping it constant. 
Therefore, our \textit{GNet} takes $E$ 
as input to the convolutional layers. 
The convolutional weights are demodulated using the encoded appearance $z$ to finally reconstruct the image $I$. 
It comprises four residual blocks and four upsample residual blocks that transform an input tensor of dimensions $H/16 \; \times \; W/16 \; \times \; 512$ to an RGB image of dimensions $H \times W \times 3$. 
The architectural design of  \textit{GNet} follows StyleGAN2 \cite{Karras2019stylegan2} including bilinear upsampling, equalised learning rate, noise injection at every layer, adjusting variance of residual blocks and leaky ReLU. 
See Fig.~\ref{fig:pipeline} for an overview of StylePoseGAN. 

\subsection{Pose-Appearance Disentanglement by Paired Training}
When the images of the same person in different poses are available for training, we can use them to further disentangle the pose and appearance. 
Given the source and target image pair $(I_s, I_t)$ of the same subject, we extract the corresponding appearances and poses $A_s, A_t, P_s$ and $P_t$. 
%
Next, we encode the poses and appearances using \textit{P-Net} and \textit{A-Net} to obtain the encodings $E_s, E_t, z_s$ and $z_t$. 
Finally, we generate the images $I_s'$ and $I_{s\rightarrow t}'$ as 
\begin{equation} 
\label{eq:paired_training} 
\begin{aligned} 
& I_s' = \text{\textit{GNet}}(E_s, z_s),\;\text{and}\\ 
& I_{s\rightarrow t}' = \text{\textit{GNet}}(E_s, z_t). 
\end{aligned} 
\end{equation} 
Here $I_{s\rightarrow t}'$ is the generated image of the source appearance in the target pose, which can be directly supervised during the training. 

\subsection{Training Details and Loss Functions}
\label{sec:training_details}
Given the input pairs $(I_s, I_t)$ and the generated images  $I_s', I_{s\rightarrow t}'$, the entire architecture is trained end-to-end for the parameters of \textit{PNet, ANet} and \textit{GNet}. 
We optimise the following loss: 
 
\begin{equation}
\label{eq:total_loss}
\begin{split}
\mathcal{L}_{total} = \mathcal{L}(I_s', I_s) + \mathcal{L}(I_{s\rightarrow t}', I_t) + \lambda_{patch}L_{patch}(I_{s\rightarrow t}', I_t).
\end{split}
\end{equation}
The total loss $\mathcal{L}_{total}$ consists of reconstruction loss $\mathcal{L}(\cdot)$ and patch co-occurrence loss $L_{patch}(\cdot)$.
The reconstruction loss 
\begin{equation} 
\mathcal{L}(I_{gen}, I_{gt}) = \lambda_{L1}L_1 + \lambda_{VGG}L_p + \lambda_{face}L_{face} + \lambda_{GAN}L_{GAN} 
\end{equation} 
comprises the following terms: 

\begin{itemize}
\item \textit{$L_1$ reconstruction loss.} We use $L_1$ distance as a reconstruction loss to force $I_{gen}$ and $I_{gt}$ to be close to each other: 
\begin{equation} 
L_1 = |I_{gen} - I_{gt}|. 
\end{equation} 
\item \textit{Perceptual Reconstruction Loss.} We use a perceptual loss \cite{johnson2016perceptual}  based on the VGG Network to enforce perceptual similarity between generated and the ground truth image. It is defined as the difference between the activations on different layers of the pre-trained VGG network \cite{simonyan2014very} applied on $I_{gen}$ and $I_{gt}$: 
\begin{equation} 
L_{VGG} = \sum \frac{1}{N^j}|p^j(I_{gen}) - p^j(I_{gt})|, 
\end{equation} 
where $p^j$ is the activation and $N^j$ the number of elements of the $j$-th layer in the VGG network pre-trained on ImageNet. 
\item \textit{Face Identity Loss.} We use a pre-trained Face Identity Network to enforce similarity of the facial identity between $I_{gen}$ and $I_{gt}$: 
\begin{equation} 
L_{face} = |N_{face}(I_{gen}) - N_{face}(I_{gt})|, 
\end{equation} 
where $N_{face}$ is the pre-trained SphereFaceNet \cite{liu2017sphereface}. 
\item \textit{Adversarial Loss.} We use an adversarial loss $L_{GAN}$ with a discriminator \textit{D} of the identical architecture as in  StyleGAN2 \cite{Karras2019stylegan2}. 
Please refer to StyleGAN2 for further details. 
\end{itemize}

In addition, we use \textit{Patch Discriminator Loss} $L_{patch}$ with a patch co-occurrence discriminator \textit{DPatch}. 
DPatch is trained 
such that the patches in $I_{gen}$ can not be distinguished from 
patches in $I_{gt}$. 
A similar idea is used by \cite{park2020swapping} in an unsupervised setting. 
Please refer to \cite{park2020swapping} for the architecture of \textit{DPatch}. 

The final objective $\mathcal{L}_{total}$ is minimised w.~r.~t.~the parameters of \textit{PNet, ANet} and \textit{GNet}, while maximised w.~r.~t.~\textit{D} and \textit{DPatch}. 

\begin{figure}[t!]
    \includegraphics[width=\linewidth]{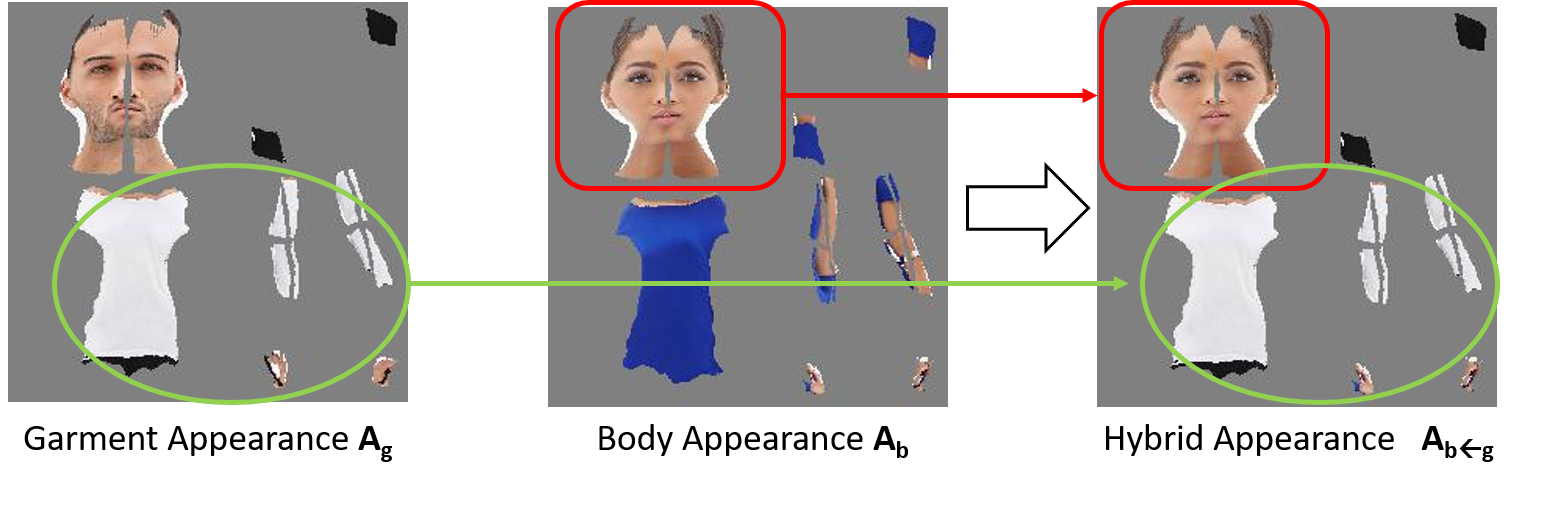}
    \caption{\textbf{Hybrid appearance for garment transfer.} Given the appearance images of the body $A_b$ and garments $A_g$, we construct a hybrid appearance image $A_{b \leftarrow g}$ with the garment specific regions in $A_g$ and body-specific regions in $A_b$.}
    \label{fig:hybrid_appearance} 
\end{figure}

\subsection{Inference} 
Once trained, our single trained model can be used for pose transfer, garment and attribute transfer, and interpolation in the learned manifold of appearance. 
In summary, we can control the pose encoding $E$ and the appearance encoding $z$ independently to accomplish a wide range of tasks. 

\subsubsection{Pose Transfer}
\label{sec:pose_transfer}
For pose transfer, our method takes a source appearance $A_s$ (that is extracted from a human image), and a target DensePose $P_t$ as input. The re-rendered image of the source person in the target pose is then obtained as 
\begin{equation}
     I_{s\rightarrow t} = \text{\textit{GNet}}(\textit{PNet}(P_t),  \textit{ANet}(A_s)). 
\end{equation} 

\subsubsection{Garments and Parts Transfer}\label{sec:garment_transfer} 
Once trained for pose transfer, our model can be used for garment transfer without any modification. 
Given a source body image $I_b$ (with appearance $A_b$) and a target garment image $I_g$ (with appearance $A_g$), the aim is to reconstruct the person in $I_b$ with the garments in $I_g$ as $I_{b \leftarrow g}$. 
To achieve this task, we construct a hybrid appearance image $A_{b \leftarrow g}$ with the garment specific regions in $A_g$ and body-specific regions in $A_b$, see Fig.~\ref{fig:hybrid_appearance}. 
The generated image with the swapped garments is then $I_{b \leftarrow g} = \text{\textit{GNet}}(\textit{PNet}(P_b), \textit{ANet}(A_{b \leftarrow g}))$. 
Here \textit{GNet}, \textit{PNet} and \textit{ANet} are the trained models for the task of pose transfer. 
\section{Experimental Results} 
\label{sec:results}

\subsection{Experimental Setup} 
We use the \textit{In-shop Clothes Retrieval} benchmark of DeepFashion dataset \cite{Liu2016DeepFashion} for our main experiments. 
The dataset comprises around $52k$ high-resolution images of fashion models with $13k$ different clothing items in different poses. 
We use the training and testing splits provided by  \cite{Siarohin2019AppearanceAP}. 
To filter non-human images, we discard all the images where we could not compute DensePose, resulting in $38k$ training images and $3k$ testing images. 
We train our system with the resulting training split and use the  testing split for conditioning poses. 
We also show the qualitative results of our method on Fashion dataset \cite{Zablotskaia2019DwNetDW} that has $500$ training and $100$ test videos, each containing roughly $350$ frames. 

We train our model for the task of pose transfer with paired data, as described in Sec.~\ref{sec:training_details}. 
In our experiments, texture resolution $H_a \times W_a$ is chosen to be $256 \times 256$, while the output resolution $H \times W$ is chosen to be $256 \times 256$ and $512 \times 512$ depending on compared methods. 
The loss weights (Sec.~\ref{sec:training_details}) are set empirically to $\lambda_{L1} = 1, \lambda_{VGG} = 1, \lambda_{face} = 1, \lambda_{GAN} = 1, \lambda_{patch} = 1$. 
For training, we use ADAM optimiser \cite{adam} with an initial learning rate of $0.002$, $\beta_1=0.0$ and $\beta_2=0.99$. 
After the convergence of the training, we use the same trained model for all the tasks, \textit{i.e.,} pose transfer, garment transfer, style interpolation and motion transfer. 

\subsection{Pose Transfer} 
\begin{figure*}[t]
    \includegraphics[width=0.81\linewidth]{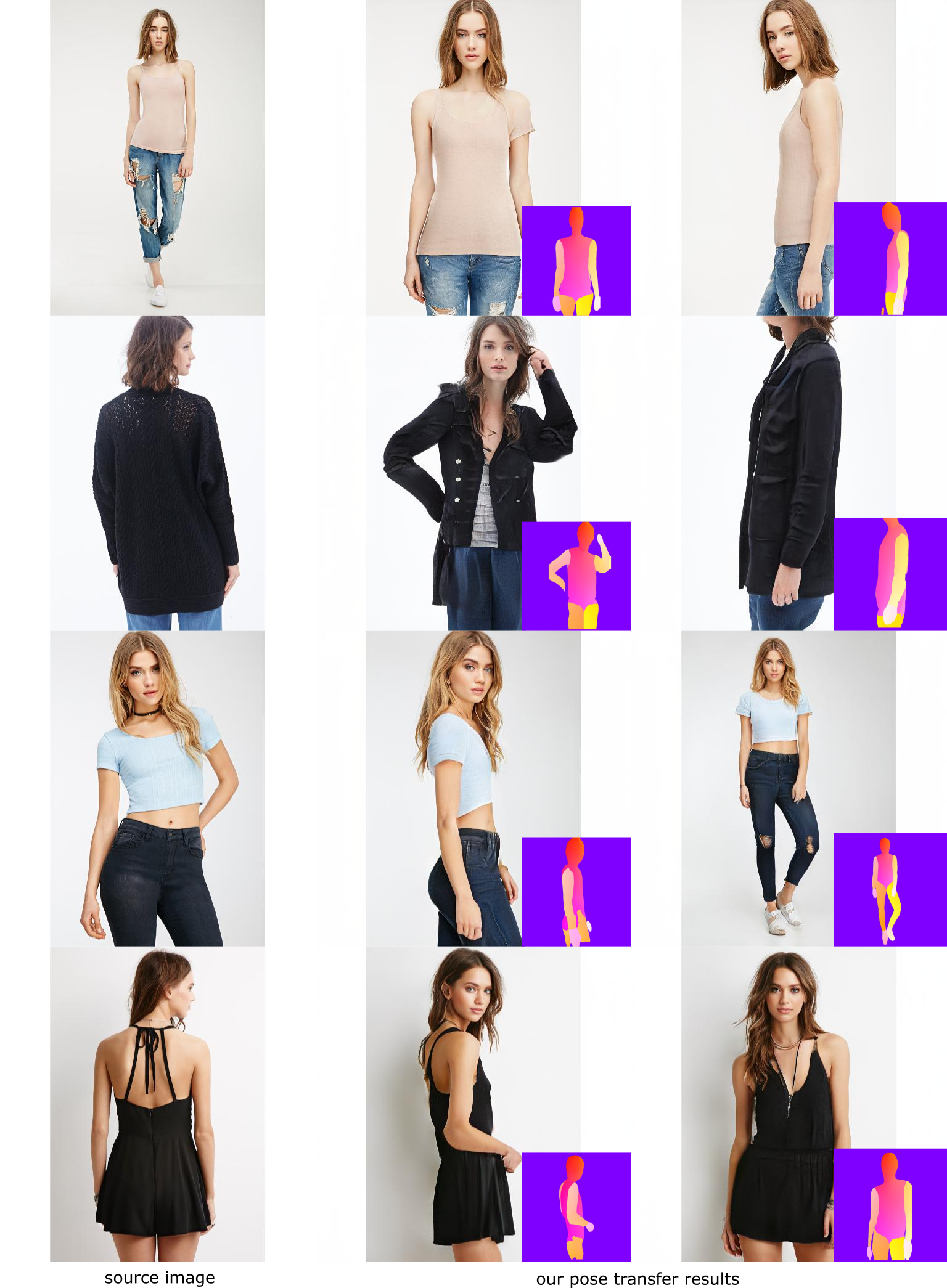}
    \caption{\textbf{High-resolution results (}$\boldsymbol{512 \times 512}$\textbf{) of our method for pose transfer}. The conditioning pose is shown on the bottom right for each generated image.} 
    \label{fig:pt_high_res} 
\end{figure*}

We perform the experiment of pose transfer using the DeepFashion dataset. 
Given a human image and a target pose, we re-render the human in the target pose as described in Sec.~\ref{sec:pose_transfer}. Our qualitative results are shown in Figs.~\ref{fig:soa_pose} and \ref{fig:pt_high_res}.

\subsubsection{Comparison with State of the Art}\label{sec:results:sota} 
\begin{figure*}[t]
    \includegraphics[width=\linewidth]{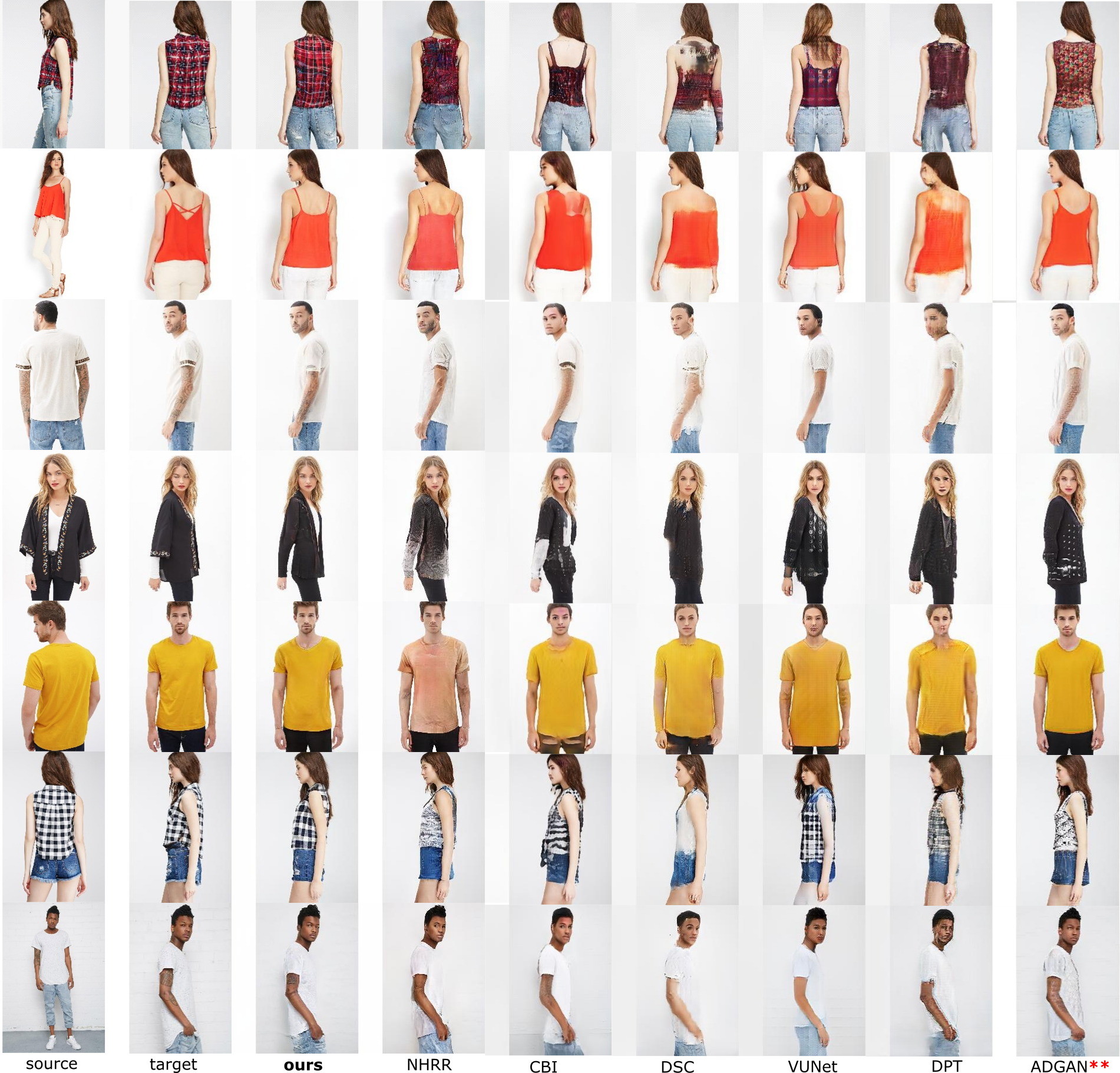}
    \caption{\textbf{Results of our method},   
    NHRR \cite{Sarkar2020},
    CBI \cite{Grigorev2019CoordinateBasedTI},
    DSC \cite{Siarohin2019AppearanceAP}, 
    VUnet \cite{esser2018variational}, 
    DPT \cite{Neverova2018},
    and ADGAN** \cite{men2020controllable}.
    Our approach produces higher-quality renderings and fine-scale details than the competing methods. \\
    **The testing pairs were \textit{included} during the training for ADGAN. See Sec.~\ref{sec:results:sota} for more details.} 
    \label{fig:soa_pose} 
\end{figure*}

We compare our results with seven state-of-the-art methods: Coordinate Based Inpainting (CBI)  \cite{Grigorev2019CoordinateBasedTI}, Deformable GAN (DSC) \cite{Siarohin2019AppearanceAP}, Variational U-Net (VUnet) \cite{esser2018variational}, Dense Pose Transfer (DPT) \cite{Neverova2018},  Neural Human Re-Rendering (NHRR) \cite{Sarkar2020}, and ADGAN \cite{men2020controllable} at the resolution $256 \times 256$ and show the qualitative results in Fig.~\ref{fig:soa_pose}. 
We train and evaluate our model  with the training-testing split  provided by Deformable GAN \cite{Siarohin2019AppearanceAP}. 
This split is also used by all the aforementioned pose transfer  methods with the exception of ADGAN \cite{men2020controllable}. 
Training the official implementation of ADGAN with our training split did not converge.  
Therefore, we provide here the results from their trained model  in-spite of the significant overlap of the testing pairs in their  training. 

It can be seen that our results show higher realism and better preserve the identity and garment details compared to the other methods. 
We observe that StylePoseGAn also faithfully reconstructs \textit{high-frequency details}, such as textures and patterns in the garments, which was not captured by any of the competing methods.
%

We next perform a quantitative evaluation with a subset of the entire testing pairs -- $176$ testing pairs that are used in the exiting works  \cite{Sarkar2020,Grigorev2019CoordinateBasedTI}. 
The following two metrics were used for comparison: 
\begin{itemize}
    \item 
\textit{Structural Similarity Index (SSIM) }\cite{ssim2004}. 
SSIM has been widely used in the existing literature for the  problem of pose transfer. 
However, this metric often does not reflect human perception. 
It is observed that smooth and blurry images tend to have better SSIM  than sharper images \cite{zhang2018perceptual,Neverova2018}. 
\item 
\textit{Learned Perceptual Image Patch Similarity (LPIPS)}  \cite{zhang2018perceptual}. 
LPIPS captures human judgment better than existing hand-designed  metrics, making it the most popular and important metric to  evaluate generated images. 
The quantitative results are shown in Table \ref{table:soa}. 
We significantly outperform the existing methods on both metrics. 
In terms of LPIPS, we observe an improvement of \textbf{19\%} (from $0.164$ to $0.133$) over the \textit{previous best result} (NHRR). 
\end{itemize} 

\begin{table}[t]
\caption{Comparison with state-of-the-art methods, using various perceptual metrics, Structural Similarity Index (SSIM) \cite{ssim2004} and Learned Perceptual Image Patch Similarity (LPIPS)  \cite{zhang2018perceptual}. $\uparrow$ ($\downarrow$) means higher (lower) is better. 
Our approach outperforms all tested methods in both metrics. 
Note that we improve LPIPS compared to the best previous method NHRR \cite{Sarkar2020} by ${\approx}20\%$. 
}
\centering
\begin{tabular}{r|r|r} 
              & SSIM $\uparrow$   & LPIPS  $\downarrow$ \\ \hline
DPT \cite{Neverova2018}           &   0.759      &     0.206  \\
VUnet  \cite{esser2018variational}       &  0.739      &    0.202   \\           
DSC \cite{Siarohin2019AppearanceAP}          &  0.750        &   0.214    \\
CBI  \cite{Grigorev2019CoordinateBasedTI}          &  0.766     &   0.178   \\
NHRR \cite{Sarkar2020} & {0.768}  &  {0.164}   \\ 
StylePoseGAN \textbf{(ours)} & \textbf{0.788}  &  \textbf{0.133}   \\ \hline
GT  & 1.0 &  0.0  \\ 
\end{tabular}
\label{table:soa}
\end{table}

\subsubsection{User Study}\label{ssec:user_study} 
\begin{figure}[t!]
    \includegraphics[width=\linewidth]{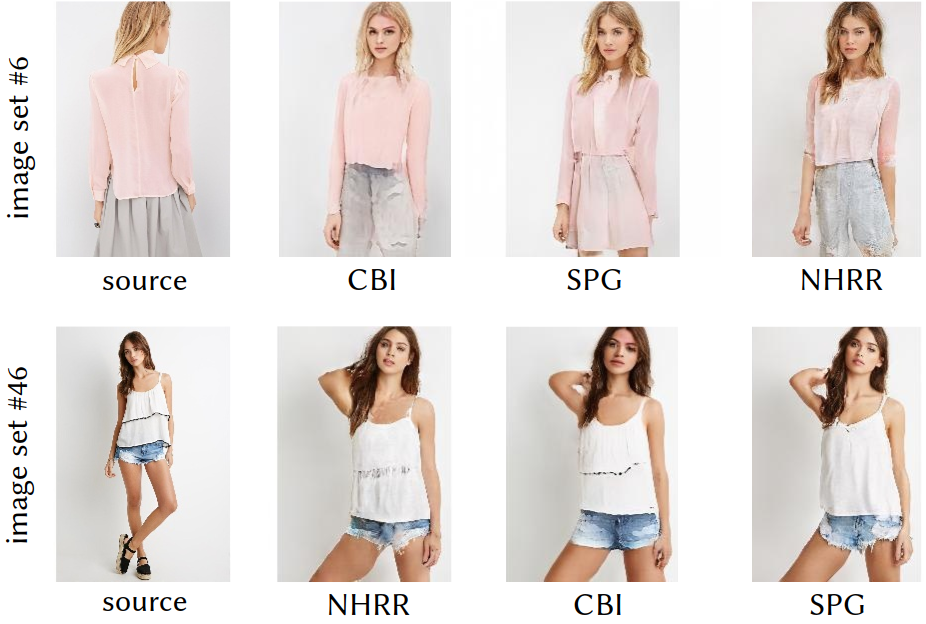} 
    \caption{Two image sets, on which CBI was preferred over other methods, \textit{i.e.,} in Q1 (``person similarity'', top row) and Q3 (``fine details'', bottom row). 
    }
    \label{fig:CBI_wins} 
\end{figure}

We conduct a user study to assess the visual quality of the pose transfer results by CBI, NHRR and our method. 
We follow the user study methodology introduced in \cite{Sarkar2020}, \textit{i.e.,} the questions cover a wide variety of source and target poses, and the distribution of males and females roughly reflects the same distribution in the training dataset. 
Moreover, multiple queries contain artefacts for our method, which make the decisions difficult. 
For each sample, we ask the following questions: 
\begin{enumerate} 
    \item[(Q1)] Which view looks the most like the person in the source image? 
    \item[(Q2)] Which view looks the most realistic? 
    \item[(Q3)] Which view preserves fine appearance details (\textit{e.g.,} texture patterns, wrinkles and other elements) better? 
\end{enumerate} 
We prepare $46$ questions asked in a web-browser user interface in a randomised order. 
Each question contains a real source image of a person and there pose transfer results (by CBI, NHRR and our method) in randomised order. 
$27$ anonymous respondents 
have submitted their answers, and the results are summarised in Table \ref{table:user_study}. 
Our method ranks first and is preferred in all three question types by a significant margin compared to  CBI and NHRR. 
CBI is preferred twice out of $144$ cases, \textit{i.e.,} once in Q1 (``person similarity'') and once in Q3 (``fine details''), see Fig.~\ref{fig:CBI_wins} for the corresponding image sets. 
Note that even if CBI was selected two times, the remaining two questions to the same set of images were decided in favour of our method (\textit{e.g.,} CBI preserved the identity best, and, at the same time, StylePoseGAN produces the most realistic rendering with fine details preserved best). 
According to the final binary per-question rankings, NHRR is not included in Table \ref{table:user_study}, however, in many cases, it was close to the most frequently voted  method (\textit{i.e.,} either CBI or ours). 
All in all, the user study shows that  StylePoseGAN lifts state of the art in pose  transfer on a new qualitative level. 
%


%

\begin{table}[t] 
\caption{The summary of the user study.} 
\label{table:user_study}  
\centering 
\begin{tabular}{r|r|r} 
                    & CBI & StylePoseGAN (ours) \\ \hline 
  Q1 (``person similarity'')    & $2.2\%$  & $97.8\%$  \\    
  Q2 (``overall realism'')      & $0\%$    & $100\%$   \\
  Q3 (``fine details'')  & $2.2\%$  & $97.8\%$  
\end{tabular} 
\end{table}

\subsection{Ablation Study} 
\label{sec:ablation}

\begin{figure*}[t]
    \includegraphics[width=\linewidth]{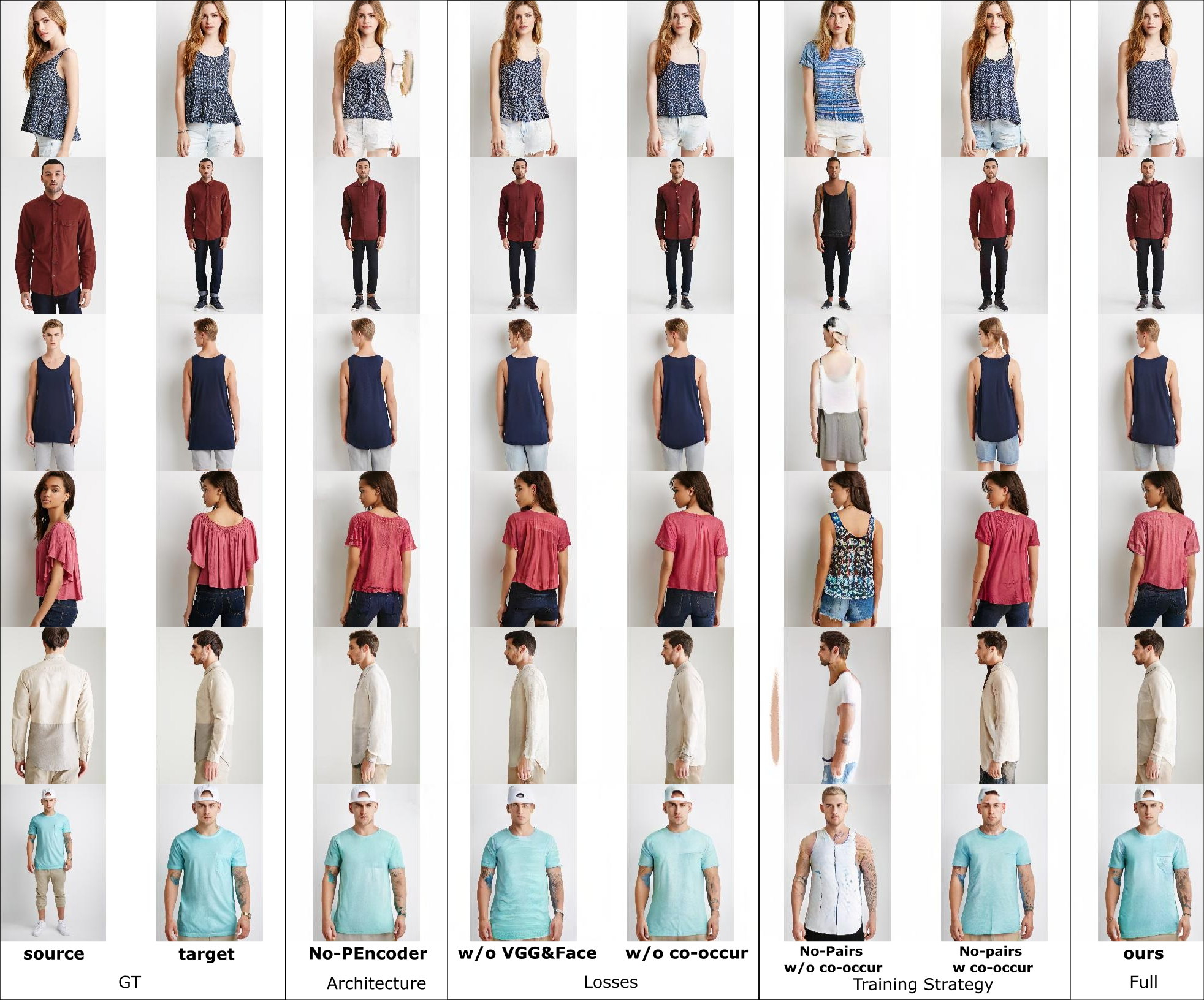}
    \caption{\textbf{Results of our ablation study.} Here, \textit{No-PEncoder} does not use pose encoder, but a generator requiring more memory; 
    \textit{w/o VGG\&Face} and \textit{w/o co-occur} do not use VGG/Face and co-occurrence loss, respectively; \textit{No-Pairs} baselines do not use paired training. 
    We observe that our full method performs better than the baselines in almost all cases. 
    The importance of co-occurrence loss during unpaired  training is discussed in Sec.~\ref{sec:ablation}.} 
    \label{fig:ablation} 
\end{figure*}

We perform the following ablation study to see the usefulness of different components. The qualitative results are shown in Fig.~\ref{fig:ablation}.

\subsubsection{Architecture} 

\begin{figure*}[t]
    \includegraphics[width=\linewidth]{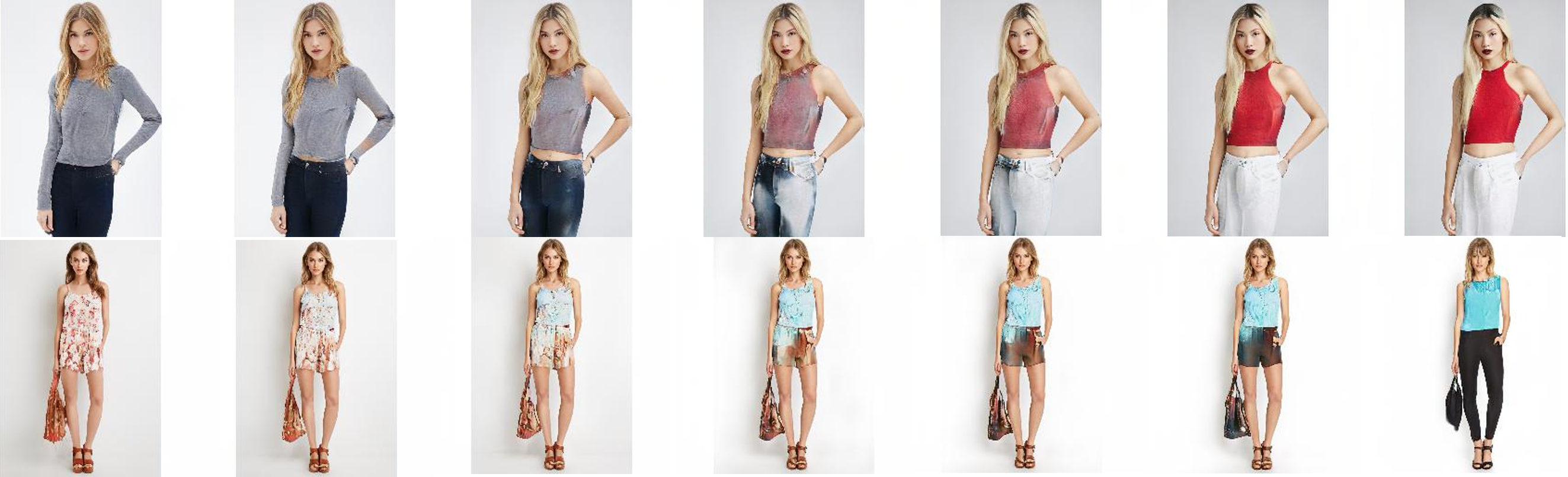}
    \caption{\textbf{Latent space interpolation.} We can interpolate between two appearance vectors to generate coherently dressed humans with the properties of both target appearances.} 
    \label{fig:latent_space} 
\end{figure*}

The spatial structure $P$ can be directly fed into the generator instead of encoding it through a separate encoder \textit{PNet}. 
However, the StyleGAN2-like generator takes an input of small spatial dimensions; resizing the pose to such small dimensions ($16 \times 16$ for the output resolution of $256 \times 256$) removes all the meaningful structure information. 
We, therefore, change \textit{GNet} with two upsample residual blocks (instead of four in our original design) to take the pose $P$ as input, and completely drop \textit{PNet} -- for the ablation experiment \textbf{No-PEncoder}. 
Because of the large memory requirements by the generator, this baseline takes longer to train. 
However, after convergence, it performs well and generates high-quality images. 
In comparison to our full method, \textit{No-PEncoder} takes four times longer to train, and is more unstable with bad input poses as shown in Fig.~\ref{fig:ablation}. 

Note that style-based generative models can be extended to their conditional version where the conditioning variable is mapped to its style input $W$ (Fig.~\ref{fig:stylegan2}) while keeping the rest of the architecture same, \textit{i.e.,} the spatial tensor $S_{input}$ is a learned constant. 
However, we could not bring such network to convergence (with the same generator architecture \textit{GNnet}) in our supervised settings. 
Both of the designs---\textit{i.e.,} mapping of the concatenated pose $P$ and appearance $A$ by an \textit{ANet} like encoder, and mapping of pose and appearance through separate convolution network followed by fully connected network---did not work. 
We hypothesise the reasons to be (a) the challenging nature of human images (in comparison to faces) where it is difficult for the mapping network to learn both the appearance and pose together in a single vector, (b) smaller size of our generator in comparison with the StyleGAN which makes it more difficult to learn in these challenging settings and (c) difficulty for a StyleGAN based generators to be extended to the pure conditional scenario. 
The later case can be mitigated by injecting noise to compromise the supervision errors. 
However, we did not conduct more exhaustive experiments in this direction, as our primary focus of this work is the spatial condition for style-based generators. 

\subsubsection{Losses} To see the usefulness of the different loss terms, we perform the experiment \textbf{w/o VGG+Face} and \textbf{w/o patch concurrency} by removing VGG and face losses, and patch co-occurrence. 
We see with both qualitative and quantitative results that VGG and face identity losses are crucial for the final results. 

\subsubsection{Training strategy} We often do not have a training dataset containing images of the same person with multiple poses. To address the issue, we propose the ablation experiment \textbf{No-Pairs} where we do not use source-target pairs for the training. Given an image $I_s$, this design reconstructs back $I_s$ by its own appearance $A_s$ and pose $I_s' = \text{\textit{GNet}}(\textit{PNet}(P_s), \textit{ANet}(A_s))$. In addition, we use a random pose from the training set $P_t$, and use it to reconstruct 
$I_{s\rightarrow t} = \text{\textit{GNet}}(\textit{PNet}(P_t), \textit{ANet}(A_s))$. However, we do not apply any reconstruction loss on $I_{s\rightarrow t}$ because of unavailability of $I_t$ during the training. Instead, we apply patch co-occurrence loss $L_{patch}$ between $I_{s\rightarrow t}$ and $I_s$. 
Therefore, the loss objective of Eq.~\eqref{eq:total_loss} is modified to 
\begin{equation} 
\mathcal{L}_{total} = \mathcal{L}(I_s', I_s) +  \lambda_{patch}L_{patch}(I_{s\rightarrow t}', I_t). 
\end{equation} 
This experiment accounts for the unsupervised version of our method. We also perform an experiment with the exact setting but with no additional patch co-occurrence loss. 
A similar method is proposed in swapping autoencoder  \cite{park2020swapping}. 
In contrast to Park et al., this experiment explicitly uses normalised pose and normalised appearance as input which enables better control of the conditioning pose and appearance. 
This experiment also provides us with the model for garment transfer and has all the advantages of our full method. 

The usefulness of the co-occurrence loss can be seen in this unpaired scenario. 
As observed in Fig.~\ref{fig:ablation},  \textbf{No-Pairs} with co-occurrence loss performs well. However, it lacks texture details and adds spurious patterns when the poses are too different. 
\textit{No-pairs} without co-occurrence loss do not preserve appearance. The paired training in our full method version forces the generator to account for the missing texture in highly occluded areas. 

\begin{figure*}[t]
    \includegraphics[width=\linewidth]{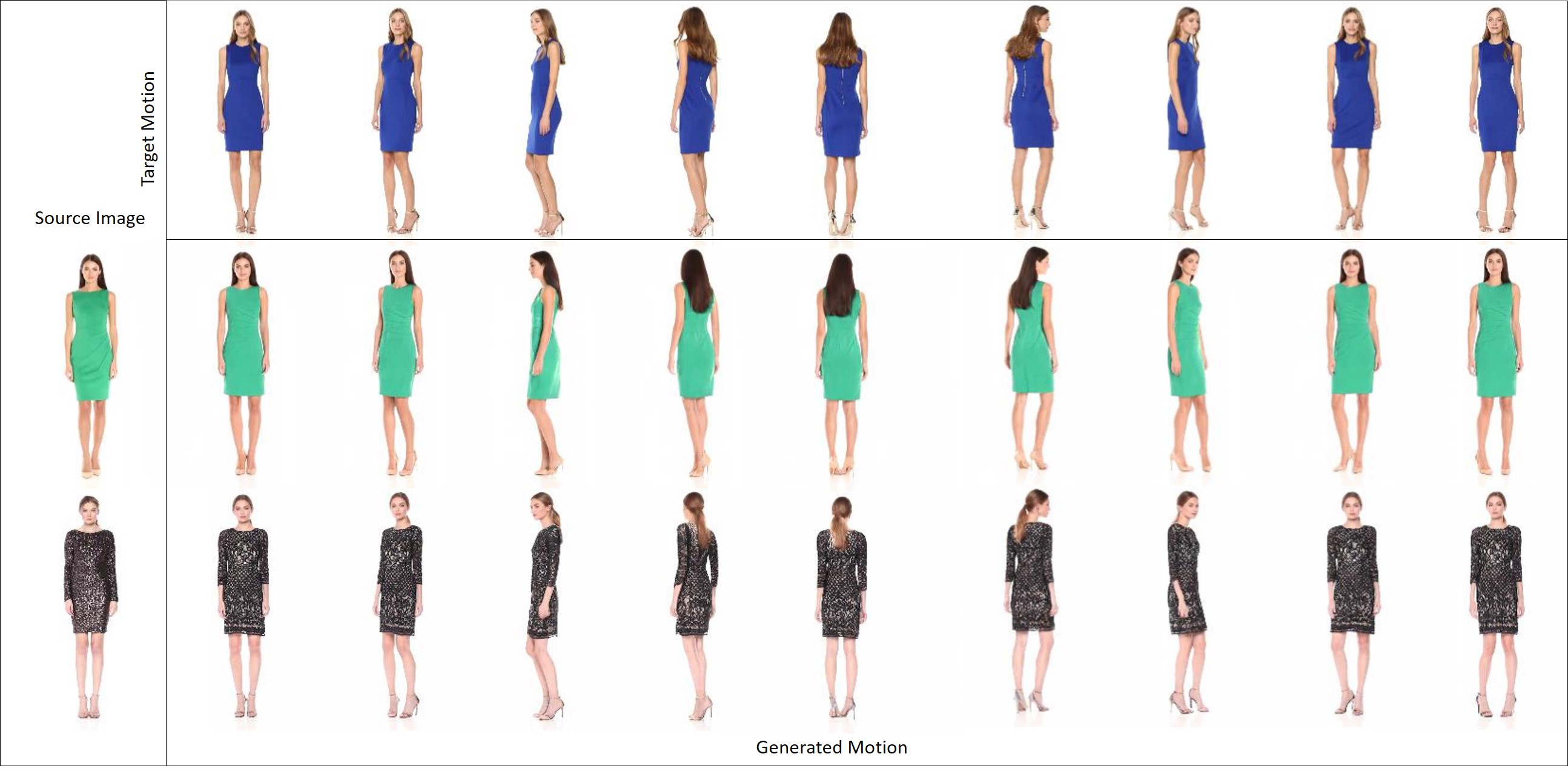}
    \caption{\textbf{Motion transfer from a single image.} StylePoseGAN can produce motion animations from a single image from the target sequence. 
    Here, we generate the frames with the pose from \textit{target motion} and appearance from \textit{Source Image}. 
    Please see the accompanying video for the  visualisation of pose-dependent appearance  changes.} 
    \label{fig:motion_transfer} 
\end{figure*}

\begin{figure*}[t]
    \includegraphics[width=\linewidth]{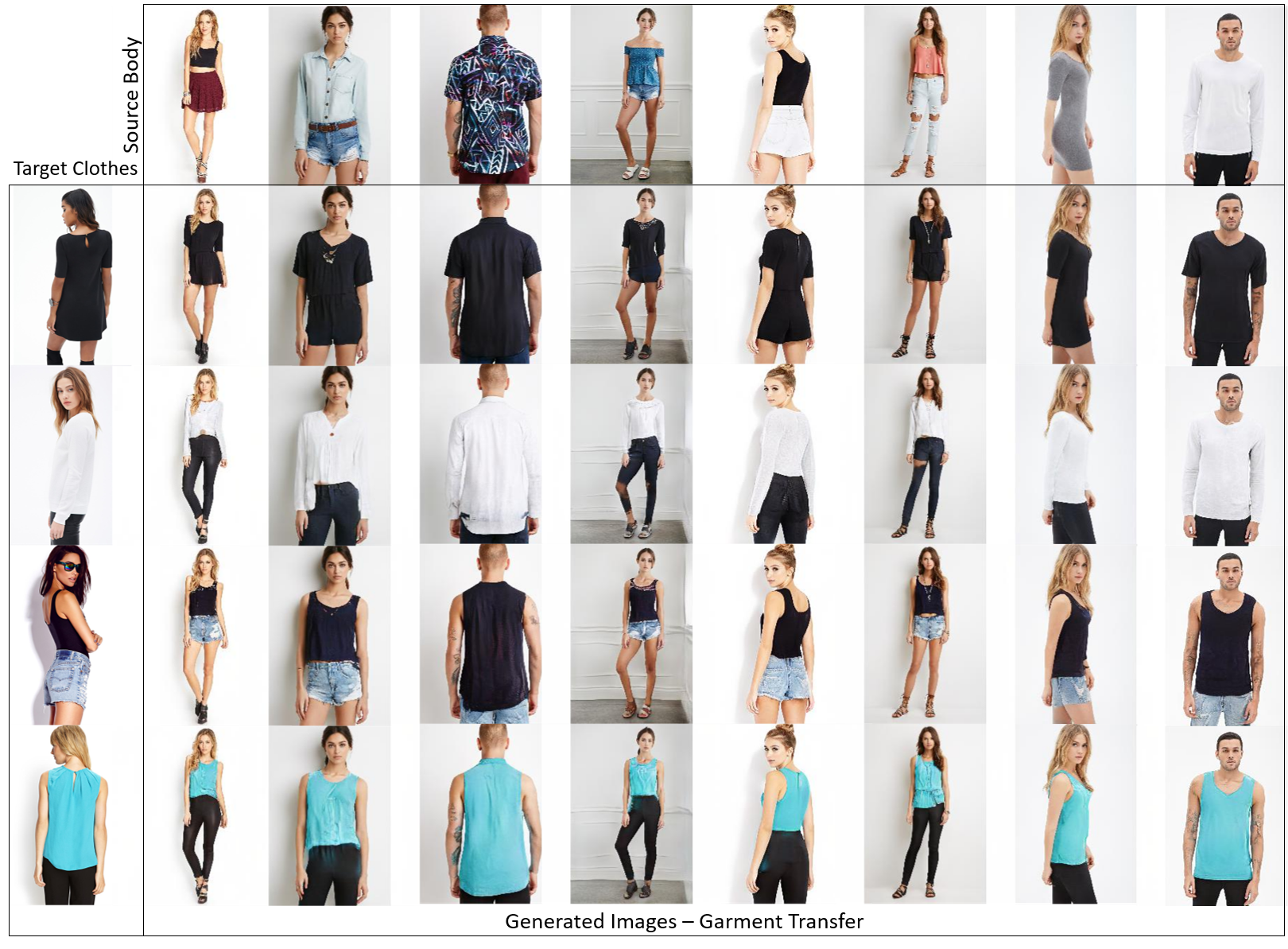}
    \caption{\textbf{Our garment transfer results.} The images are generated with the bodies in the row \textit{source body} and garments in the column \textit{target clothes}. 
    Best viewed with zoom. 
    } 
    \label{fig:garment_transfer} 
\end{figure*}

\begin{figure*}[t]
    \includegraphics[width=\linewidth]{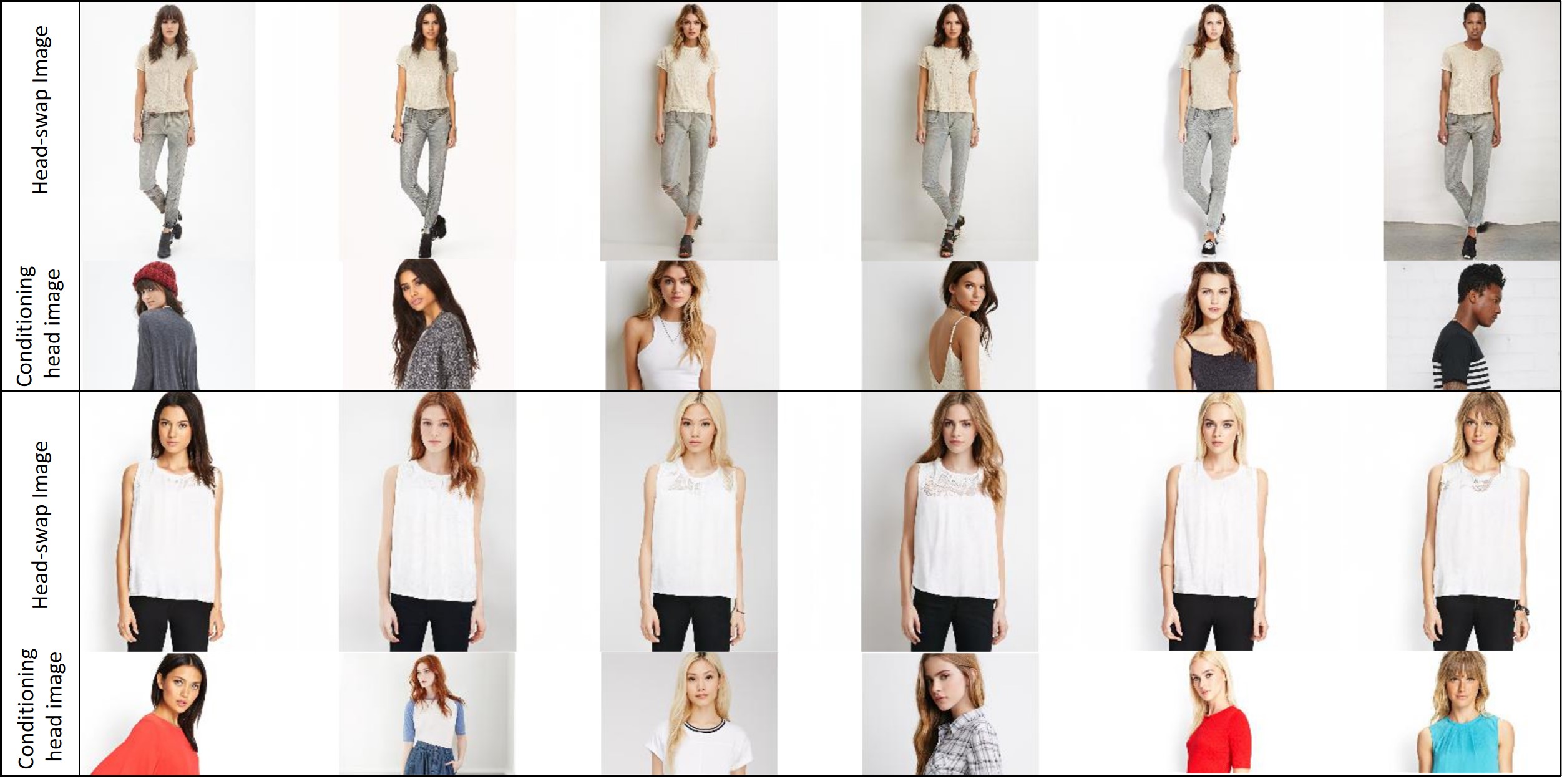}
    \caption{\textbf{Our results of \textit{head-swap}.} The conditioning image of the head is shown on the bottom for each generated image.} 
    \label{fig:head_swap} 
\end{figure*}

\subsection{Motion Transfer} 
%
Our method can be applied to each frame of a driving video to perform motion transfer. 
To this end, we keep the image of the source person fixed and use the pose from the actor of the driving video in our system to create image animation. 
Note that we do not exploit any temporal context, and we create the video frame by frame. 
We show our result on Fashion Dataset \cite{Zablotskaia2019DwNetDW} in Fig.~\ref{fig:motion_transfer} and our supplementary video. 
We observe that our method does not show the shower-curtain effect that is typically present in translation-network-based generators (such as Pix2Pix). 
Even with the errors and inconsistencies in DensePose on the  testing frames, our method keeps the fine wrinkle patterns  consistent and provides pose-dependent appearance changes.
Note that these wrinkle patterns are learned implicitly by our method through the training data. 

\subsection{Garment and Part Transfer}
As explained in Sec.~\ref{sec:garment_transfer}, our part-based encoding enables our model to perform garment transfer without 
any further training, \textit{i.e.,} no explicit dataset of garments are needed for our garment transfer functionality. 
The results are shown in Fig.~\ref{fig:garment_transfer}. 
We observe that StylePoseGAN faithfully reconstructs the garments and body details, and seamlessly generates coherent human images with swapped garments. 

The same idea can be used to transfer any other body parts. To this end, we show an example of \textit{head-swap} in Fig.~\ref{fig:head_swap}. 
Here we use the hybrid appearance $A$ that contains the partial texture of body from one image and head from another image, and the pose from the body image to create the head swap effect. 
Note that the identity is well preserved even though some  conditioning head images are side views. 

\subsection{Style Interpolation} 
We find that our latent space is smooth. 
%
Interpolating appearance features $z_{inter}$ according to 
\begin{equation} 
  z_{inter} = z_1 t + z_2(1-t),\;t \in [0, 1], 
\end{equation} 
where $z_1, z_2$ are the appearance encodings of two human images, results in images of coherently dressed humans with the properties of both images. 
Examples are shown in Fig.~\ref{fig:latent_space}. 
\section{Discussion}
\label{sec:discuss}

\begin{figure}[t]
    \includegraphics[width=\linewidth]{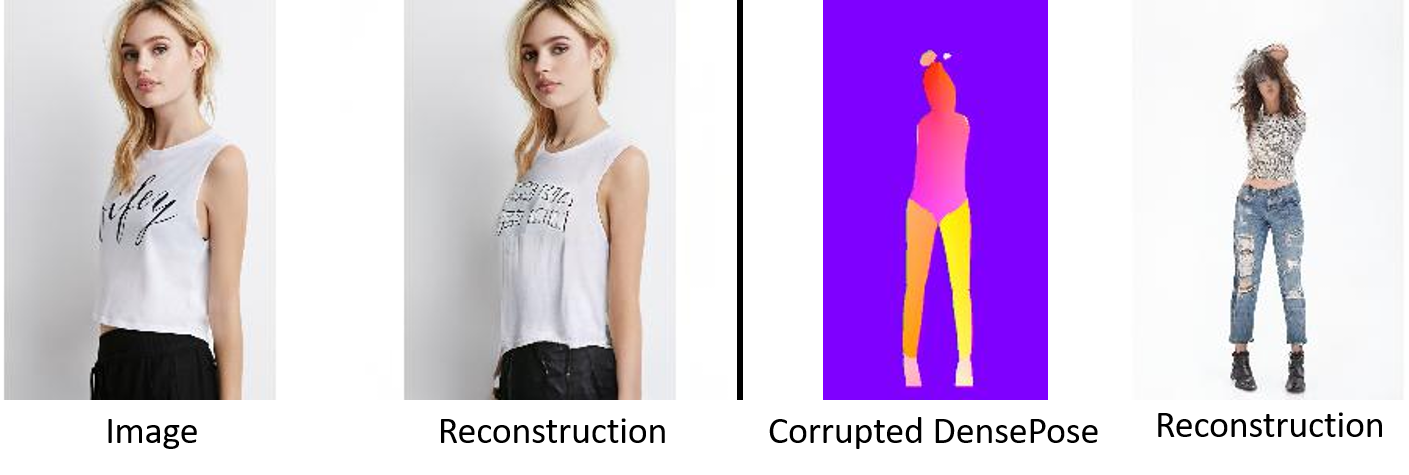}
    \caption{\textbf{Failure cases.} \textit{(Left)}  Encoding the appearance with a vector makes our  method difficult to reconstruct highly entangled  spatial features such as text. 
    \textit{(Right)} Bad conditioning DensePose image  during the test time deteriorates the output. 
    Also, see Fig.~\ref{fig:challangeing_cases} for  challenging cases.} 
    \label{fig:failure_cases} 
\end{figure}

\begin{figure*}[t]
    \includegraphics[width=\linewidth]{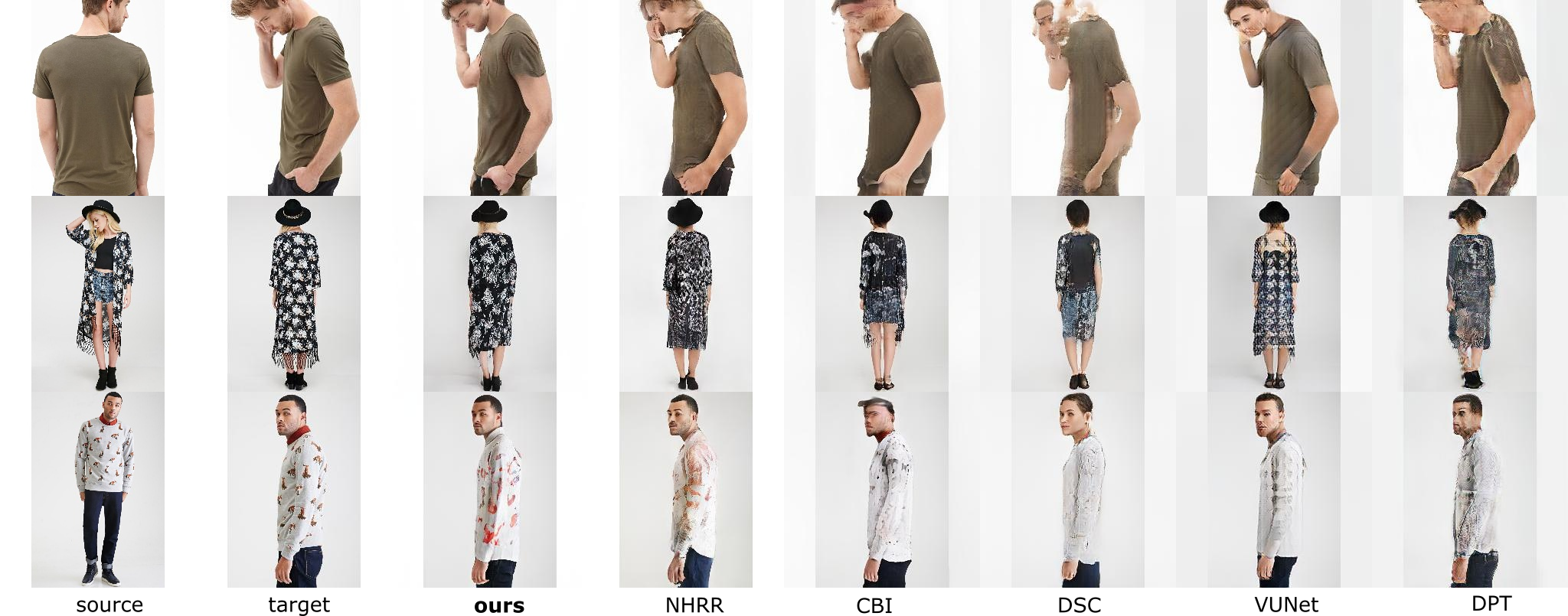}
    \caption{\textbf{Highly challenging cases.} Even  though our method struggles under 
    extreme occlusions, target poses and textures, 
    it synthesises more plausible images than the compared methods. 
    } 
    \label{fig:challangeing_cases} 
\end{figure*}

\paragraph{Limitations/Failure Cases}
 Because we encode the appearance by fully connected layers to a single vector, the spatial semantics in the appearance is destroyed. 
 Therefore, our method has difficulty in reconstructing highly entangled spatial features such as images or text in the clothes (see Figs.~\ref{fig:failure_cases} and  \ref{fig:challangeing_cases}). 
 Even in those cases, our method performs better than translation-network-based methods such as NHRR because of the powerful nature of our generator.

\section{Conclusion} 
We presented StylePoseGAN, \textit{i.e.,} a new method for synthesising  photo-realistic novel views of humans from a single monocular image allowing for explicit control over the pose and appearance without the need for sophisticated 3D modelling. 
StylePoseGAN significantly outperforms the current state of the art in the perceptual metrics and achieves a high level of realism of the synthesised images in our experiments. 
The quantitative results are confirmed by a comprehensive user study, in which our results are preferred over competing methods in $142$ cases out of $144$. 
We conclude that this is due to improved support of fine texture details of the human appearance such as facial features, textures and shading. 
The ablative study has confirmed that all design choices are necessary for the best photo-realistic results. 
We thus believe that our method, which often produces deceptively realistic renderings, is an important step towards unconstrained novel view rendering of scenes with humans, opening up multiple avenues for future research. 
\label{sec:conclusion}

\bibliographystyle{ACM-Reference-Format}
\bibliography{article.bib}
\end{document}